\pdfoutput=1
\documentclass{article}
\usepackage{ijcai20}

\usepackage{times}
\usepackage{soul}
\usepackage{url}
\usepackage[hidelinks]{hyperref}
\usepackage[utf8]{inputenc}
\usepackage[small]{caption}
\usepackage{graphicx}
\usepackage{amsmath,bm}
\usepackage{amsthm}
\usepackage{booktabs}
\usepackage{algorithm}
\usepackage{algorithmic}
\usepackage{array}
\usepackage{latexsym}
\usepackage{booktabs}
\usepackage{subfigure}
\usepackage{multirow}
\usepackage{makecell}
\usepackage{CJKutf8}
\usepackage{color}
\usepackage[normalem]{ulem}

\def\newcite#1{\citeauthor{#1}~\shortcite{#1}}

\title{Multi-Task Learning with Auxiliary Speaker Identification for Conversational Emotion Recognition}

\author{
Jingye Li$^1$
\and
Meishan Zhang$^2$\footnote{Corresponding author}\and
Donghong Ji$^1$\And
Yijiang Liu$^1$
\affiliations
$^1$School of Cyber Science and Engineering, Wuhan University, China\\
$^2$School of New Media and Communication,Tianjin University, China
\emails
\{theodorelee, dhji, cslyj\}@whu.edu.cn,
mason.zms@gmail.com,
}

\begin{document}
\begin{CJK}{UTF8}{gbsn}
\maketitle

\begin{abstract}
Conversational emotion recognition (CER) has attracted increasing interests in the natural language processing (NLP) community.
Different from the vanilla emotion recognition, effective speaker-sensitive utterance representation is one major challenge for CER.   
In this paper, we exploit speaker identification (SI) as an auxiliary task to enhance the utterance representation in conversations.
By this method, we can learn better speaker-aware contextual representations from the additional SI corpus.
Experiments on two benchmark datasets demonstrate that the proposed architecture is highly effective for CER, obtaining new state-of-the-art results on two datasets.
\end{abstract}

\section{Introduction}
Emotion recognition has been one hot topic in natural language processing (NLP) which aims to 
detect emotions in texts \cite{wen2014emotion,li2015sentence}. 
Recently, emotion recognition in conversions has been received increasing attentions \cite{majumder2019dialoguernn,zhong2019knowledge,ghosal2019dialoguegcn}.
Given a sequence of utterances by multiple speakers, conversational emotion recognition (CER)
aims to recognize emotion for each utterance.
CER is a typical sequence labeling problem, and end-to-end neural sequence labeling models 
have achieved state-of-the-art performance \cite{poria2017context,Jiao2019higru}.  

Intuitively, speaker information can be greatly helpful for CER. 
For example, the last utterance of a same speaker could be severed as an important clue for the current utterance. 
Thus, how to effectively represent the speaker-sensitive utterances in conversions is critical for CER models.
Previous studies, e.g., ConGCN \cite{zhang2019Modeling} and DialogueGCN \cite{ghosal2019dialoguegcn},
build graphical structures over the input utterance sequences by speaker information and then exploit graph neural networks 
to model the dependencies, leading to better performance for CER.  

\begin{figure}[tb]
  \centering
  \includegraphics[width = 0.48 \textwidth]{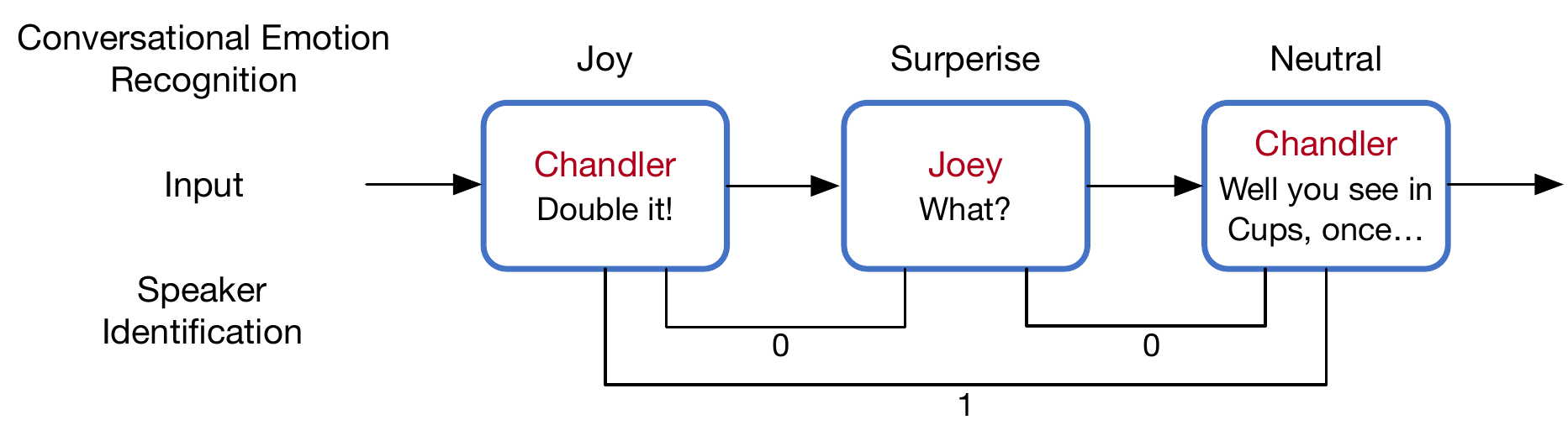}
  \caption{Illustration of multi-task learning of our method. For CER, the emotion of every utterance in conversation is predicted. For SI, several pairs of utterances will be selected for binary classification, where $1$ denotes the same speaker and $0$ otherwise. }
  \label{figure_example}
\end{figure}

The above models just adopt speakers as indicators to build connections over sequential utterances,
using only CER training corpus to learn speaker-aware representations,
which could be insufficient for speaker exploration.
In most cases, we can have much larger scale corpora of raw conversions, where no emotion information is annotated.
These corpora could be potentially useful to learn speaker-aware contextual representations,
since the utterances as well as the corresponding speaker identities are offered jointly in them. 
Speaker identification (SI) could be one good alternative for this purpose.
As shown in Figure \ref{figure_example}, 
we can learn speaker-aware contextual representations through the raw conversions by judging 
whether two utterances are from the same user.


In this work, we propose to use SI as one auxiliary task in order to obtain better speaker-aware contextual representations of the conversational utterances.
We exploit a multi-task learning (MTL) framework to achieve our final goal to enhance CER.
For CER and SI, we use the same network structure for utterance encoding,
but with different set of model parameters.
We adopt BERT as the basic representation to make our baseline strong,
and hierarchical bidirectional gated recurrent neural networks (Bi-GRU) are exploited 
at the utterance-level and conversation-level to enhance the contextual representations.
Further, we unite the two tasks  with two bridging network structures by using an attention network \cite{bahdanau2014neural} and a gate mechanism \cite{wu2019different}  at the utterance-level and conversation-level for full mutual interaction, respectively . 


We conduct experiments on two benchmark datasets to verify our framework, 
which are respectively EmoryNLP and MELD by name, 
both are sourced from the TV show of Friends.
The results show that our baseline system is highly strong,
achieving better performance than previous state-of-the-arts. 
Our final model can lead to significant improvements on the two datasets both,
which demonstrates the effectiveness of our proposed method.
In addition, we conduct extensive analysis work to examine the model in depth,
for better understanding the advantages of our model. 
All codes and experimental settings will be released publicly available on https://github.com/ThdLee/CER for research purpose under Apache License 2.0.


\section{Related Work}

Emotion recognition in conversational texts is generally treated as a sequence labelling problem in the literature. 
Traditional approaches often use lexicon-based and acoustic features to detect emotions \cite{riley2004predicting,devillers2006real}. 
Recently, deep learning based recurrent neural networks (RNN) and transformer can bring state-of-the-art performance \cite{poria2017context,tzirakis2017end,zhong2019knowledge},
which is able to capture contextual utterance representations effectively. Our baseline CER model follow these settings, 
adopting a sophisticated RNN for contextualized representation learning.

Several studies attempt to integrate speaker information in the conversational utterances, as it can affect the final CER performance much \cite{Hazarika2018conversational,Hazarika2018icon}. 
\newcite{majumder2019dialoguernn} propose a recurrent model to detect emotion by tracking party state and global state dynamically. 
Graph convolutional networks (GCN) have been demonstrated stronger in modeling context-sensitive and speaker-sensitive dependence in conversation \cite{zhang2019Modeling,ghosal2019dialoguegcn}.
In this work, we propose an improved approach to better 
utilize speaker information by multi-task learning with a closely-related auxiliary task.

Multi-task learning aims to model multiple relevant tasks
simultaneously, which is capable of exploiting potential shared features across tasks effectively. 
There have been a number of successful studies in the NLP community \cite{liu2017adversarial,ma2018modeling,xiao2018gated,pentyala2019multi,wu2019different}. 
\newcite{zhou2019multi} exploit language modeling as an auxiliary task to assist question generation task by multi-task learning, which motivates our work greatly. Similar to the language modeling task,
speaker identification has rich training corpus, which can be collected automatically,
and we exploit it to assist CER mainly. 


\section{Methodology}
\begin{figure}[t]
  \centering
  \includegraphics[width=0.95\linewidth]{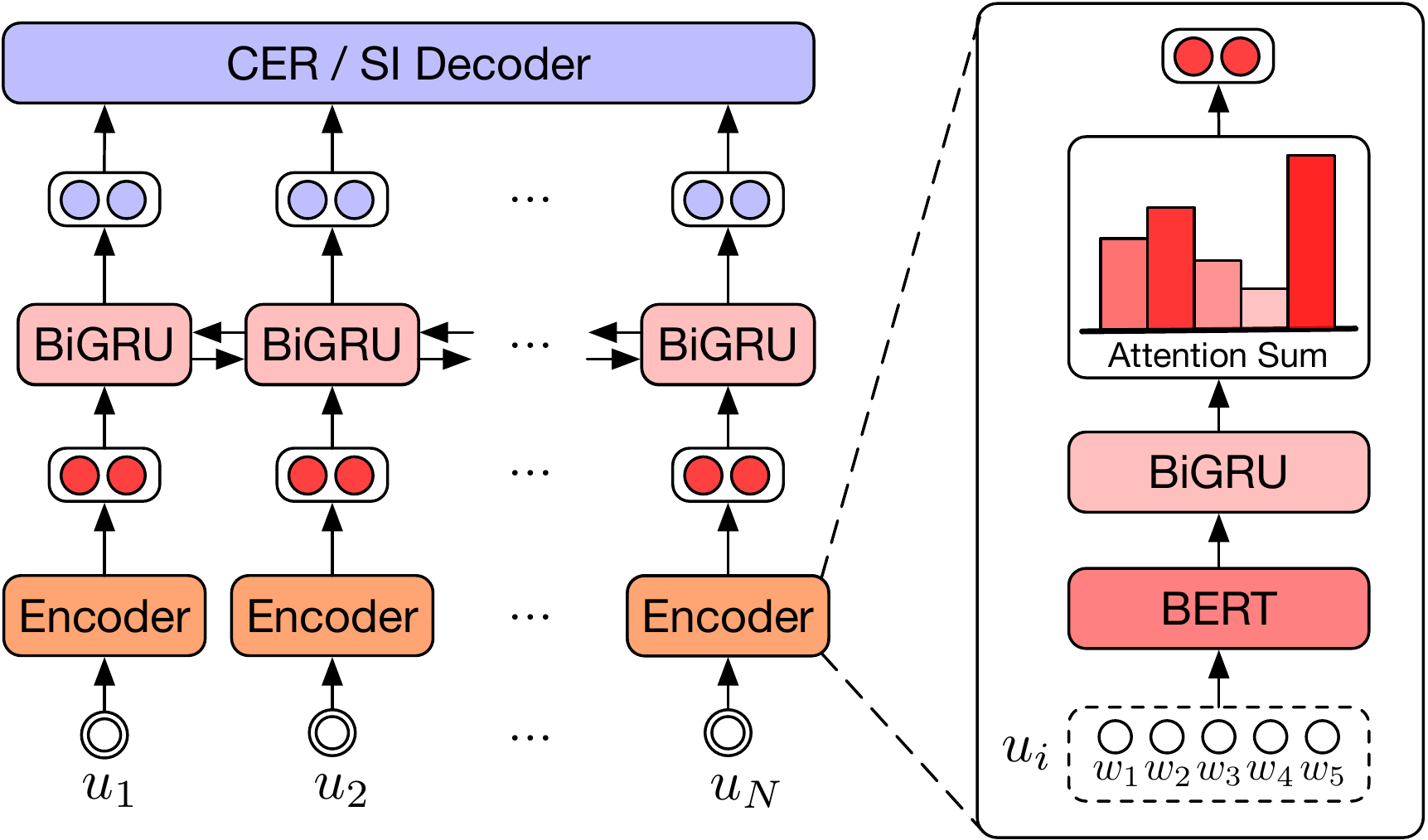}
  \caption{The overall architecture of our individual models.}
  \label{figure_single_task}
\end{figure}

\begin{figure*}[t]
  \centering
  \includegraphics[width=0.95\linewidth]{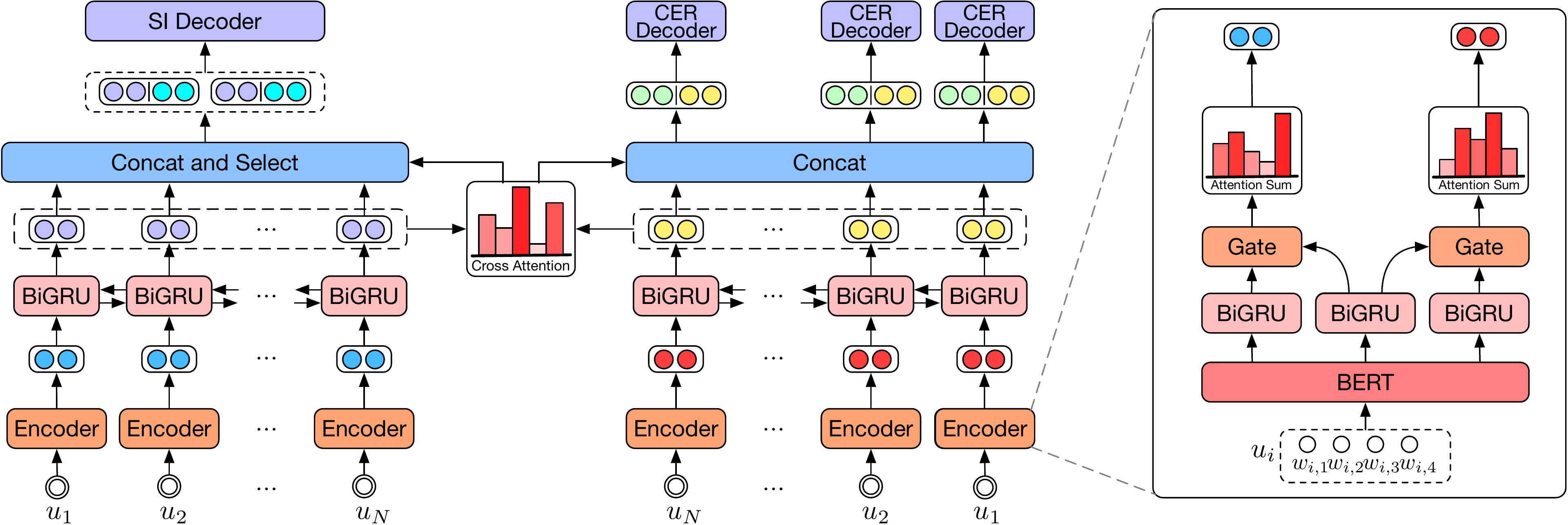}
  \caption{Illustration of our proposed model for CER and SI.}
  \label{figure_multi_task}
\end{figure*}

Suppose we have a conversation with $N$ consecutive utterances ${u_1, u_2, \cdots, u_N}$ and ${M}$ speakers ${p_1, p_2, \cdots, p_M}$. Each utterance ${u_i}$ is uttered by one speaker ${p_{S(u_i)}}$, where the function ${S}$ maps the index of the utterance into its corresponding speaker. 
The objective of CER is to predict the emotion label of each utterance, and the objective the auxiliary task SI is to classify whether two given utterances ${u_p, u_q}$ in a conversation are from the same speaker.
In this section, we will introduce our specific-task model and multi-task learning framework in detail. 
Figure \ref{figure_single_task} shows the architecture of our baseline model, which exploits attention-based hierarchical network as encoder.
Figure \ref{figure_multi_task} depicts the multi-task learning framework of our proposed model. 

\subsection{Conversational Emotion Recognition}
For the baseline CER model, first a hierarchy encoder is built on the input conversion utterance sequence, resulting in one feature vector for each utterance,
then a standard emotion classification is performed based on the encoder sequence.
Our hierarchy encoder consists of two components: individual utterance encoder and contextual utterance encoder, 
where the individual utterance encoder is an attention-based network  with Bi-GRU, 
as shown by the right part of Figure \ref{figure_single_task},
and the contextual utterance encoder is a Bi-GRU over the output sequence of individual utterance encoder,
aiming to capture the conversation-level context.\footnote{Bi-GRU is used as the basic RNN operation here by considering both efficiency and effectiveness. }

\paragraph{Individual Utterance Encoder}
The input of our model is a sequence of utterances ${u_1, \cdots, u_N}$. 
Assuming that the $i$th utterance $u_i$ is denoted by ${w_{i,1}, \cdots, w_{i,L(i)}}$, where $L(i)$ is the length of $u_i$.
We exploit BERT as the basic encoder because it has achieved state-of-the-art performances for a number of NLP tasks \cite{Devlin2019bert}, obtaining the word-level output for $u_i$: $\{ \bm{x}_{i,1}, \cdots, \bm{x}_{i,L(i)} \} $. 

Further, we adopt a single-layer Bi-GRU to further enhance the contextualized word representation at the utterance level,
which can be formulated as: 
\begin{equation}
\{ \bm{h}_{i,1}, \cdots, \bm{h}_{i, L(i)} \} = \text{Bi-GRU}^{w}(\bm{x}_{i,1},  \cdots, \bm{x}_{i, L(i)})
\end{equation}
where $\bm{h}_{i,*}$ denotes the Bi-GRU output for word $w_{i,*}$. 

The goal of individual utterance encoder is to derive a single feature vector for each utterance based on the covered words.
Next, we need to aggregate the word-level outputs into a single vector for the utterance.
We exploit an attention-based aggregation to accomplish the goal.
Formally, 
the utterance representation is defined as follows:
\begin{equation}
\begin{split}
    &\bm{z}_{i,j} = \mathrm{tanh} (\bm{W}_a \bm{h}_{i,j} + \bm{b}_a ) \label{eq1} \\
    &\alpha_{i,j} = \frac{ \mathrm{exp} (\bm{v}_a^\top \bm{z}_{i,j})}{\sum\nolimits_{k} \mathrm{exp}(\bm{v}_a^\top \bm{z}_{i,k})} \\
    &\bm{u}_i = {\sum_{j}} \alpha_{i,j} \bm{h}_{i,j}, 
\end{split}    
\end{equation}
where $\bm{W}_a$, $\bm{b}_a$ and $\bm{v}_a$ are model parameters, $\top$ indicates  vector transposition, $\bm{u}_i^e$ is the vector representation for utterance $u_i$.
Intuitively, $\alpha_{i,j}$ is the importance of word $w_{i,i}$ in $u_i$,
and the weighted sum is adopted for the utterance representation.  


\paragraph{Contextual Utterance Encoder}
After the individual utterance encoder, we obtain a sequence of utterance-level representations $\{\bm{u}_1, \cdots,  \bm{u}_N  \}$.
These representations are solely sourced from their internal words.
In order to encode utterance-level contextualized information,
we build a second Bi-GRU as follows:
\begin{equation}
\{ \bm{f}_i, \cdots, \bm{f}_N \} = \text{Bi-GRU}^{u}(\bm{u}_1, \cdots,  \bm{u}_N),
\end{equation}
where $\{\bm{f}_1, \cdots, \bm{f}_N \}$ is the final utterance presentations for prediction,
which can capture surrounding contextual information in conversations. 


\paragraph{Output Layer}
When the final contextualized utterance feature representation $\{\bm{f}_1, \cdots, \bm{f}_N \}$ is ready, we can calculate the output probability of each candidate emotion labels by a linear transformation followed with a softmax operation:
\begin{equation}
  \bm{p}^{\text{CER}}_i = \mathrm{softmax}(\bm{W}_{\text{CER}} \bm{f}_i), \text{~~s.t.~~}  i \in [1, N]
  \label{eq3}
\end{equation}
where $\bm{W}_{\text{CER}}$ is one model parameter, and $\bm{p}^{\text{CER}}_i$ denotes the output distribution of emotion labels for utterance $u_i$.

\paragraph{Training}
We optimize the CER model by minimizing the cross-entropy between the predicted emotion distribution and the true distribution. For a single conversation, the obejctive function is defined as follows:
\begin{equation}
  \mathcal{L}_{\text{CER}} = -\frac{1}{N}\sum^N_{i=1}\sum^K_{j} \bm{y}^{\text{CER}}_{i,j}\log\bm{p}^{\text{CER}}_{i,j},
\end{equation}
where $N$ and $K$ are the number of utterances in a conversation and emotion labels, respectively, $\bm{y}^{\text{CER}}_{i}$ is the one-hot vector of the ground truth emotion for utterance $u_i$, and $\bm{p}^{\text{CER}}_i$ is the predicted distribution.

\subsection{Speaker Identification}

As discussed before, speaker information plays an important role in CER.
Here we use the same hierarchical network as CER to encoder utterances for SI.
Similarly, we obtain the first stage utterance representations $\{ \bm{u}'_1,  \cdots, \bm{u}'_N \}$ by individual utterance encoder,
and then achieve the second stage utterance representations $\{ \bm{f}'_1,  \cdots, \bm{f}'_N \}$ by contextual utterance encoder.

The goal of SI is to determine whether two selected utterances are from the same speaker, which is a binary classification problem. 
To reach this goal, we randomly sample $T$ pairs of utterances for classification. 
Note that we do not extract all utterance pairs in a conversion for a balance with CER.
In addition, we do not really recognize the utterance speakers, as the classification category could be extremely large in some real scenarios, 
which makes the training very difficult.

\paragraph{Classification}
Given a sampled pair of utterance representations ${\bm{f}'_{i}, \bm{f}'_{j}}$ from a single conversation, we adopt four sources of features for SI classification: (1)$\bm{f}'_{i}$, (2) $\bm{f}'_{j}$,  (3) $\parallel \bm{f}'_{i} - \bm{f}'_{j} \parallel $,  and (4) $\bm{f}'_{i} \odot \bm{f}'_{j}$ ($\odot$ denotes element-wise multiplication).  We concatenate them, apply a nonlinear MLP layer to reach the final feature vector $\bm{f}^{\text{SI}}_{i,j}$, and then perform binary classification based on it. 
The overall process can be formalized as follows:
\begin{equation}
\begin{split}
  & \bm{\delta}_{i,j} = \bm{f}'_{i} \oplus \bm{f}'_{j} \oplus (\parallel \bm{f}'_{i} - \bm{f}'_{j} \parallel) \oplus (\bm{f}'_{i} \odot \bm{f}'_{j}) \\
  &\bm{f}^{\text{SI}}_{i,j} = \mathrm{ReLU}(\bm{W}_{f} \bm{\delta}_{i,j} + \bm{b}_{f}) \\
  &\bm{p}^{\text{SI}}_{i,j} = \mathrm{softmax}(\bm{W}_{\text{SI}} \bm{f}^{\text{SI}}_{i,j})
\end{split}
\end{equation}
where $\bm{W}_{f}$, $\bm{W}_{\text{SI}}$ and $\bm{b}_{f}$ are model parameters, $\oplus$ denotes vector concatenation,
and $\bm{p}^{\text{SI}}_{i,j}$ is a two-dimensional vector with one dim indicates the probability of the same speaker and the other dim denotes the  probability of different speakers.

\paragraph{Training}
For SI, we also adopt the cross-entropy loss between the ground truth and the predicted distribution as the training objective:
\begin{equation}
  \mathcal{L}_{\text{SI}} = -\frac{1}{T} \sum^T_{t=1}\sum_{k=0}^{1}\bm{y}^{\text{SI}}_{i,j,k} \log\bm{p}^{\text{SI}}_{i,j,k} ,
\end{equation}
where $\bm{y}^{\text{SI}}_{i,j}$ is a two-dimensional vector for the ground-truth answer, and $T$ is one hyperparameter. We randomly sample $T$ times for the utterance pair $(u_i,u_j)$.

\subsection{Multi-Task Learning}
The two above models for CER and SI have the same encoder network structure,
which brings convenience in unite them under a multi-task learning framework.
In this work, we keep the task-specific encoders, and exploit two bridging network structures for mutual interaction of the hierarchical encoders,
where one network is designed for the utterance-level individual utterance encoder,
and the other is targeted to the conversation-level contextual  utterance  encoder.


\paragraph{Individual Utterance Encoder} 
For the utterance-level individual utterance encoder, we exploit a shared-private structure to connect the two tasks.
Concretely, as shown by the right part of Figure \ref{figure_multi_task},
we add a shared Bi-GRU module to unite the two tasks.
Given the BERT outputs $\{ \bm{x}_{i,1}, \cdots, \bm{x}_{i,L(i)} \} $ ($\bm{h}_{i,j}$ ($i\in[1, N]$ and $j\in[1, L(i)]$),
we compute shared hidden vectors by Bi-GRU first:
\begin{equation}
\{ \bm{h}^{\text{sh}}_{i,1}, \cdots, \bm{h}^{\text{sh}}_{i, L(i)} \} = \text{Bi-GRU}^{\text{sh}}(\bm{x}_{i,1},  \cdots, \bm{x}_{i, L(i)}),
\end{equation}
and then design a gated mechanism to dynamically incorporate shared feature $\bm{h}^{sh}_{i,j}$ into task-specific features.
The network is mostly inspired by \newcite{wu2019different}.
The updated contextual word representations are computed as follows:
\begin{equation}
\begin{split}
    &\bm{g}_{i,j} = \sigma(\bm{W}_g\bm{h}_{i,j} + \bm{b}_g) \\
    &\hat{\bm{h}}_{i,j} = \bm{g}_{i,j} \odot \bm{h}^{\text{sh}}_{i,j}  + (1 - \bm{g}_{i,j}) \odot \bm{h}_{i,j}
\end{split}
\end{equation}
where $\bm{g}_{i,j}$ is a gate to control the portion of information flowing from the  shared Bi-GRU layer, $\sigma$ is a sigmoid function.
For the SI part, we only need to substitute $\bm{h}_{i,j}$ into $\bm{h}'_{i,j}$,
and $\hat{\bm{h}}_{i,j}$ changes to $\hat{\bm{h}'}_{i,j}$ correspondingly.
Finally, we use  $\hat{\bm{h}}_{i,j}$ and $\hat{\bm{h}'}_{i,j}$ as the individual utterance encoder outputs for CER and SI, respectively.



\paragraph{Contextual  Utterance  Encoder}
For the  union of contextual  utterance  encoder,
we adopt a cross attention mechanism to augment the utterance representation from the task of the apart side.
Taken the target CER model as an example,
we obtain one kind of extra features from the SI contextual utterance encoder. 
Concretely, assuming that the contextual utterance representations of  CER and SI 
are $\{ \bm{f}_1,  \cdots, \bm{f}_N \}$ and $\{ \bm{f}'_1,  \cdots, \bm{f}'_N \}$, respectively,
then we compute the additional feature for $\bm{f}_i$ by the following equations:
\begin{equation}
\begin{split}
  &\bm{c}_{i,j} = \bm{f}_i \bm{W}_c  \bm{f}'_j  \\
  &\beta_{i,j} = \frac{\mathrm{exp}(\bm{c}_{i,j})}{\sum\nolimits_k \mathrm{exp}(\bm{c}_{i,k})}\\
  &\hat{\bm{f}}_i = \bm{f}_i \oplus (\sum_j \beta_{i,j} \bm{f}'_{j}),
\end{split}
\end{equation}
where $\bm{W}_c$ is one model parameter. 
The attention mechanism is mainly motivated by \newcite{bahdanau2014neural} for feature selection from the apart side.
When the SI model is the target task, the mechanism is just performed at an opposite direction, and $\hat{\bm{f}'}_i$ is obtained.
Finally, we use  $\{ \hat{\bm{f}}_1,  \cdots, \hat{\bm{f}}_N \}$ and $\{ \hat{\bm{f}'}_1,  \cdots, \hat{\bm{f}'}_N \}$ instead for CER and SI decoding.




\paragraph{Multi-Task Training}
For the multi-task learning of the two tasks, we simply sum the losses of the two individual tasks together as the joint objective:
\begin{equation}
  \mathcal{L}_{\text{Multi}} = \mathcal{L}_{\text{CER}} + \mathcal{L}_{\text{SI}}
\end{equation}

\section{Experiments}

\begin{table}[t]
\centering
\resizebox{0.48\textwidth}{1.3cm}{
\begin{tabular}{l|c|c|c}
\toprule[1pt]
\multirow{2}{*}{dataset} & \#conversations & \#utterances & \multirow{2}{*}{\#avg. length} \\ \cline{2-3}
                         & train/val/test & train/val/test & \\
\midrule[0.5pt]
EmoryNLP                 & 659/89/79 & 7551/954/984 & 11.5 \\
\midrule[0.5pt]
MELD                     & 1028/114/280 & 9989/1109/2610 & 9.6  \\
\midrule[0.5pt]
\midrule[0.5pt]
Friends                  & 3329 & 61038 & 18.3 \\
\bottomrule[1pt]
\end{tabular}}
\caption{Statistics of the datasets for emotion recognition in conversation. Friends is an external dataset for MTL.}
\label{table1}
\end{table}

\subsection{Datasets}
We evaluate our model on two benchmark datasets, MELD and EmoryNLP, following previous work \cite{zhong2019knowledge}. Table \ref{table1} shows the corpus statistics.

\paragraph{MELD} \cite{poria2019meld} This is a multimodal dataset collected from TV show of Friends. There are seven emotion categories for each utterance, including anger, sadness, disgust, surprise, fear, joy and neutral.

\paragraph{EmoryNLP} \cite{Zahiri2018emotion} This dataset is also collected from TV show scripts of Friends. The difference lies in the type of emotion labels, which includes neutral, joyful, peaceful, powerful, scared, mad and sad.

In particular, we collect a large corpus for SI training, which is the entire scripts of the TV show of Friends, a superset of MELD and EmoryNLP.

\subsection{Settings}
We adopt Adam as the optimizer with the batch size of 4 to train our models, where the learning rates to fine-tune BERT and the other parameters are $1e-6$ and $2.5e-4$, respectively. Dropout rate with 0.5 is applied to avoid overfitting.
The dimension sizes of the hidden states of all the BiGRUs is set to 200 on EmoryNLP and 150 on MELD.
For evaluation, we exploit the standard weighted macro-F1 score as the major metric to measure all models, following \newcite{zhong2019knowledge}.

\subsection{Models}

For a comprehensive evaluation, we compare our model with the following baselines as well:

\paragraph{CNN} \cite{kim2014convolutional} A convolutional neural network for utterance-level classification without using contextual information at the conversation level.

\paragraph{c-LSTM} \cite{poria2017context} A hierarchical classification model based on LSTM-RNN model, where contextual utterance-level features are adopted.

\paragraph{DialogueRNN} \cite{majumder2019dialoguernn} A sophisticated RNN-based model based on three GRUs, which are used to model speakers, global contexts and historical emotions.

\paragraph{KET} \cite{zhong2019knowledge} The state-of-the-art model in the literature which exploits external commonsense knowledge to enhance the contextual utterance representation. 

\paragraph{DialogueGCN} \cite{ghosal2019dialoguegcn} A GCN-based model aiming to better representing  inter-speaker dependence, where GRU is used as the basic feature composition modules.

\paragraph{ConGCN} \cite{zhang2019Modeling} A multi-modal model for CER, which also exploits GCN to model the context-sensitive and speaker-sensitive dependence. 

\subsection{Developmental Results}
We conduct  experiments on the developmental datasets of EmoryNLP and MELD to examine our proposed models.

\begin{figure}[t]
\centering
\includegraphics[width = 0.40 \textwidth]{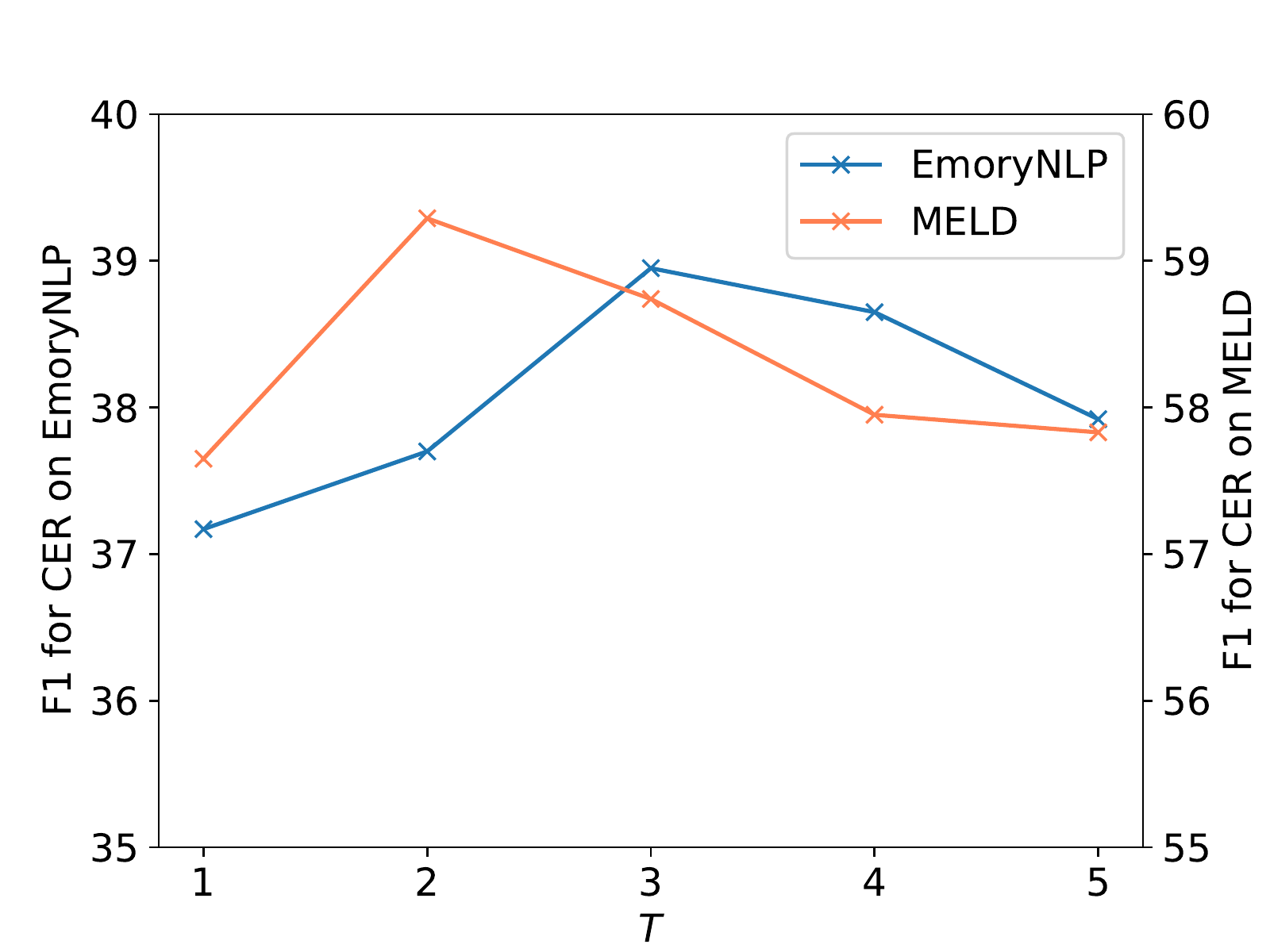}
	\caption{Developmental experimental results of our model on EmoryNLP by varying the values of $T$.}
\label{figure_T}
\end{figure}



\paragraph{The Influence of $T$ }
$T$ represents the number of utterance pairs sampled from a conversation for SI,
aiming to leverage the combined loss of the two tasks.
We investigate the influence of $T$ by ranging it from 1 to 5. 
Figure \ref{figure_T} shows the results.
As shown, we can obtain the best results when $T = 3$ on EmoryNLP and $T = 2$ on MELD, respectively,
demonstrating the importance of sampling.
In addition, the optimum $T$ of different datasets may vary, which could be possibly due to the different averaged conversation length.




\begin{figure}[t]
\begin{center}
	\subfigure[EmoryNLP]{\label{figure_finetune_emprynlp}
		\centering{\includegraphics[scale=0.31]{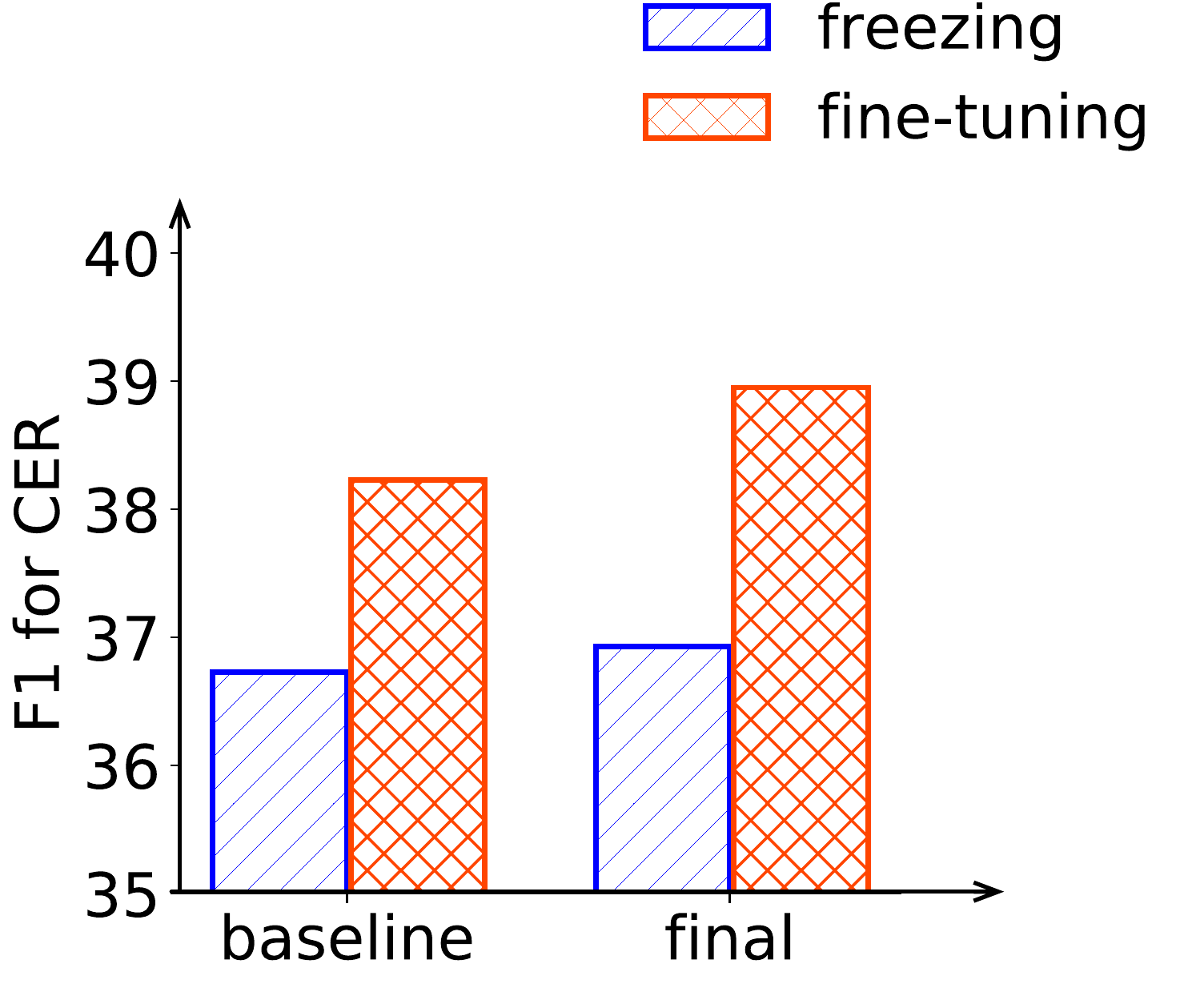}}
	}
\hspace{-1.4cm}
	\subfigure[MELD]{\label{figure_finetune_meld}
		\centering{\includegraphics[scale=0.31]{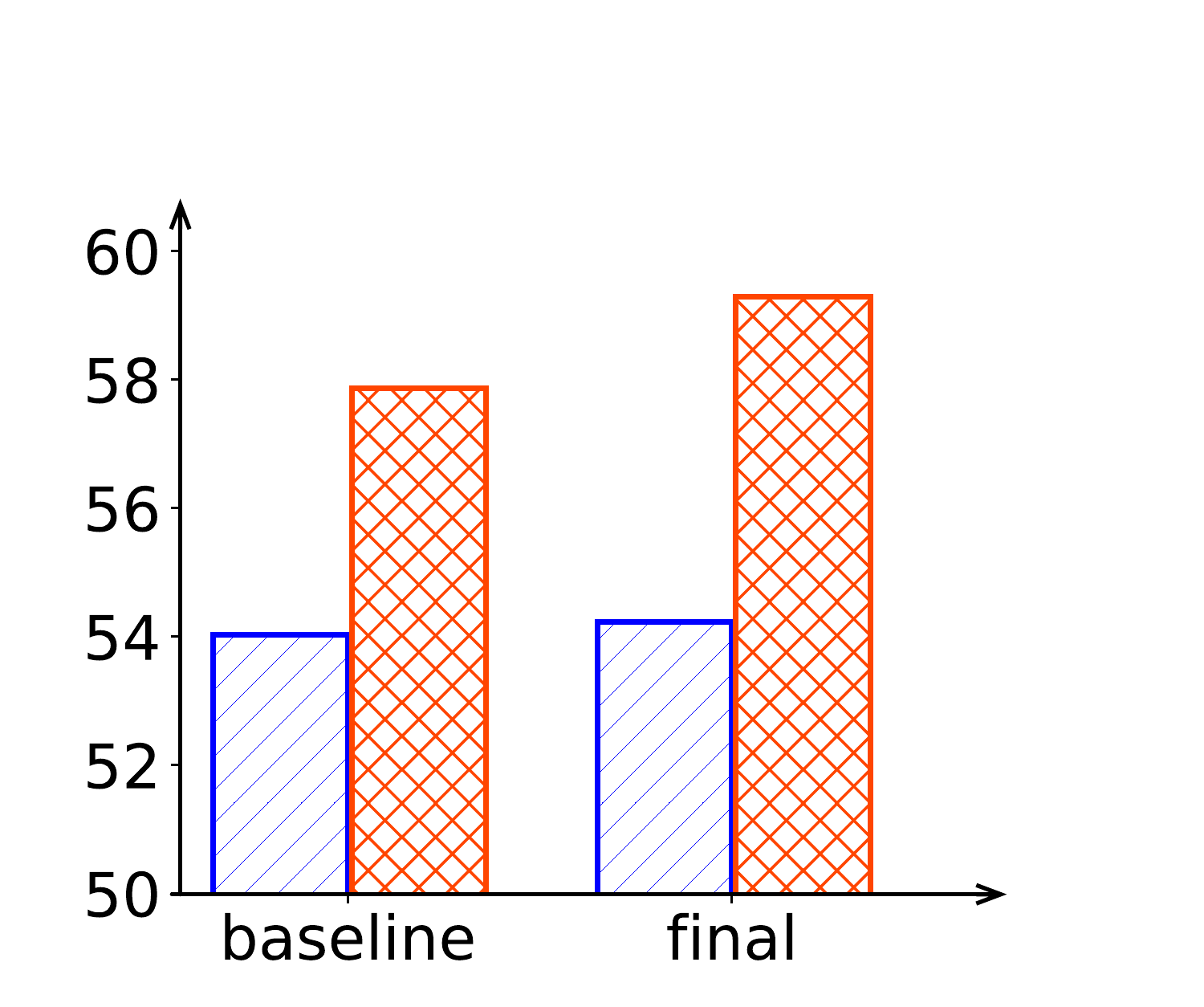}}
	}
\caption{The influence of BERT Fine-Tuning.}\label{figure_finetune}
\end{center}
\end{figure}

\paragraph{The Influence of BERT Fine-Tuning}
The utilizing of BERT should be carefully studied.
When BERT parameters are frozen, we can save the resource cost greatly, 
for example, the memory of GPU.
However, it may lead to significant performance decrease.
Here we study the gap between BERT fine-tuning and frozen.
Figure \ref{figure_finetune} shows the comparison results,
where the baseline model and our final model are both investigated.
As shown, we can see that by freezing the BERT parameters,
drops of over 2\% and 4\% can be resulted on EmoryNLP and MELD, respectively,
demonstrating that fine-tuning is a necessary for CER.

\begin{table}[t]
  \centering
  \begin{tabular}{p{1.6cm}<{\centering}|p{1.6cm}<{\centering}|p{1.6cm}<{\centering}|p{1.6cm}<{\centering}}
  \toprule[1pt]
  \multicolumn{2}{c|}{ \texttt{Interaction Mode}} & \multirow{2}{*}{EmoryNLP}  & \multirow{2}{*}{MELD} \\
  \cline{1-2}
  IUE & CUE  &  &  \\ \midrule[0.5pt]
  $\times$ & $\times$ & 38.23 & 57.86 \\
  $\times$ & $\surd$ & 38.76 & 58.79 \\
  $\surd$ & $\times$ & 38.50 & 58.49 \\
  $\surd$ & $\surd$ & \textbf{38.95} & \textbf{59.29} \\
  \bottomrule[1pt]
  \end{tabular}
  \caption{Ablated performance on EmoryNLP and MELD, where IUE and CUE denote individual and contextual utterance encoders, respectively.}
  \label{table_ablation}
\end{table}

\paragraph{Ablation Study}
To comprehensively study the effectiveness of the two bridging neural network structures at the different levels for mutual interaction, we conduct ablation experiments in detail.  
The results are offered in Table \ref{table_ablation}.
We can see that both the two networks can bring improved performance for CER.
By excluding the bridging network at IUE, our final model falls by 0.19\% on EmoryNLP and 0.50\% on MELD, respectively. Similarly, the model shows 0.45\% and 0.80\% declines on the two datasets by eliminating the bridging network at CUE, respectively. 
When both network structures are removed,  drops by 0.72\% and 1.43\% points on the two datasets are resulted, respectively.

\begin{figure*}[tb]
  \centering
  \includegraphics[width = 0.9 \textwidth]{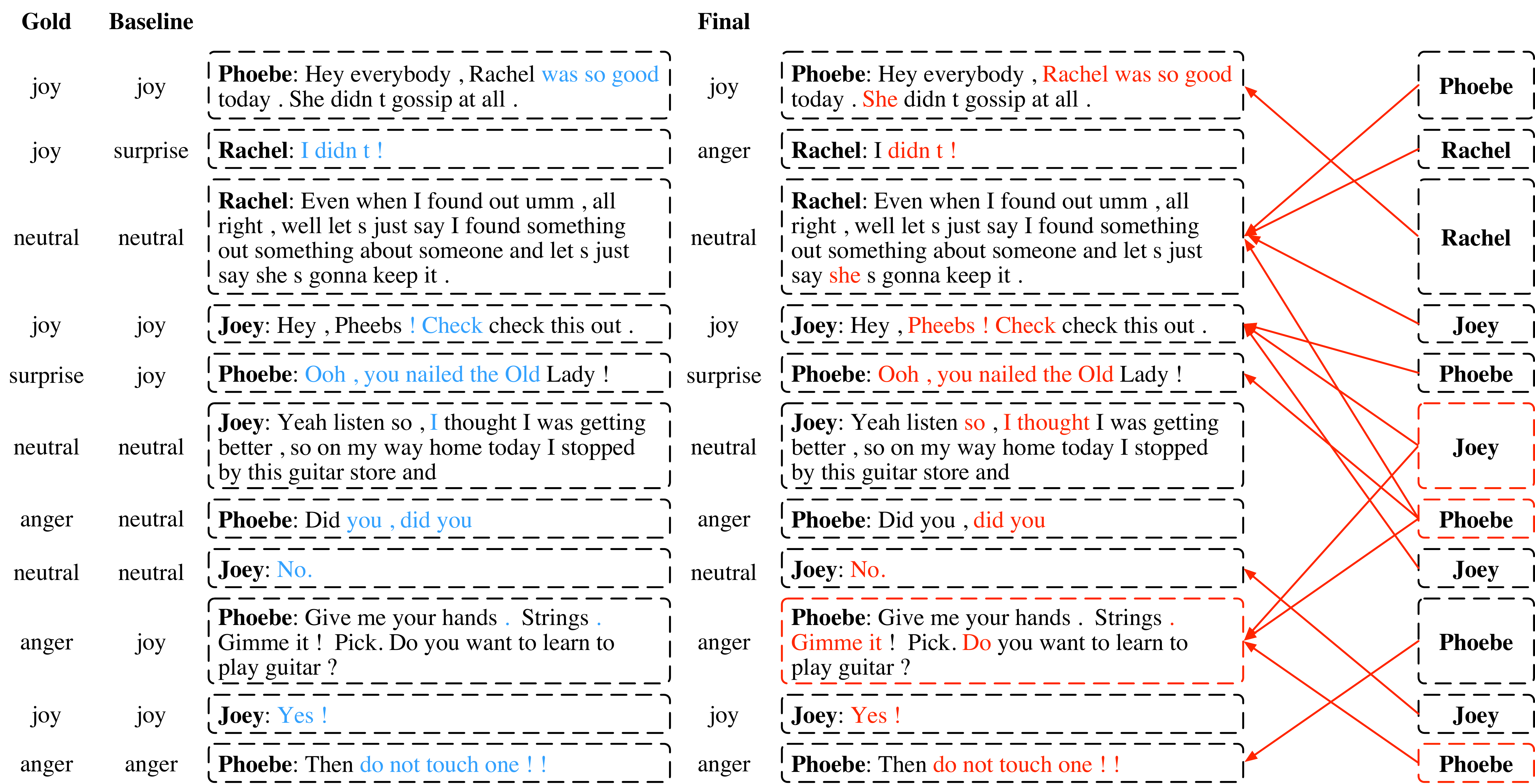}
  \caption{Visualization of the attentions, where the thresholds are 0.1 and 0.2 for words and  utterances by their attention values, respectively.}
  \label{figure_attention}
\end{figure*}



\subsection{Final Results}

\begin{table}[tb]
  \centering
  \begin{tabular}{p{2.2cm}|p{2.0cm}<{\centering}|p{2.0cm}<{\centering}}
  \toprule[1pt]
  Method                          & EmoryNLP & MELD\\
  \midrule[0.5pt]
  Our (GloVe) & 32.59 & 59.67 \\
  ~~~~~~~~+MTL & \bf 34.54 & \bf 60.69 \\ \midrule[0.5pt]
  Our (ELMo)             & 33.55 & 61.10 \\
  ~~~~~~~~+MTL & \bf 34.85 & \bf 61.86 \\ \midrule[0.5pt]
  Our (BERT)             & 34.76 & 61.31 \\
  ~~~~~~~~+MTL & \textbf{35.92} & \textbf{61.90}    \\
  \midrule[0.5pt]
  \midrule[0.5pt]
  CNN & 32.59  & 55.02 \\
  cLSTM  & 32.89 & 56.44\\
  DialogueRNN & 31.70 & 57.03 \\
  DialogueGCN & - & 58.10\\
  KET & \bf 34.39 & 58.18 \\
  ConGCN & - & \bf 59.40 \\ 
  \bottomrule[1pt]
  \end{tabular}
  \caption{Final results on the test datasets of EmoryNLP, MELD.}
  \label{table_final_results}
\end{table}

Table \ref{table_final_results} shows the performance of various models on the test sections of the two datasets, respectively. 
We report the performance of our baseline model with three kinds of word representations, 
namely the pretrained glove word embeddings \cite{Pennington2014glove} \footnote{http://nlp.stanford.edu/data/glove.6B.zip, 300d}, ELMO \cite{peters2018deep}\footnote{https://allennlp.org/elmo, original, 5.5B} and BERT\footnote{https://github.com/google-research/bert, BERT-Base, Uncased}.
We can see that all our baseline results are strong and can achieve comparable performance with the previous state-of-the-art systems.
The best-reported numbers on the EmoryNLP and MELD datasets are 34.39\% and 59.40\%, respectively.
Our BERT-based baseline gives F-scores of 34.76\% and 61.31\%, respectively,
which are both higher than the previous state-of-the-arts. 

According to the results, our models with MTL can achieve better performance on two datasets as compared to their corresponding baselines. 
On the two datasets, the F1-score improvements based on the pretrained golve embeddings are 1.95\% and 1.02\%, respectively. 
When the contextualized ELMO representations are exploited, the improvements over the baseline are 1.30\% and 0.76\%, respectively.
For our baseline based on BERT, 
the final MTL enhanced model leads to F1 increases of 1.16\% and 0.59\% on the two datasets, respectively.
All the improvements by MTL are significant (p-value below 0.0001 by using pair-wised t-test).
Interestingly, we can also find that the improvements become smaller as the baseline becomes stronger.

\subsection{Visualization Analysis}
For comprehensive understanding of our proposed models, we visualize the attention matrices by a case study, which is selected from the MELD test dataset. 
Figure \ref{figure_attention} shows the example, where both salient words of individual utterance encoder and the key utterances of cross attention neural structure both are offered.

First, we examine the difference in salient words for individual utterance encoder.
As shown, the final model treats \textit{gimme it}, \textit{she}, \textit{Rachel}  and \textit{Pheebs} as strong clues for CER,
which are speaker related, 
while misses the punctuation such as comma and period,
which are mostly objective.
The observation indicates that the shared Bi-GRU module can help to identify speaker-related words such as speaker names and attributes, highlighting them for further feature representation,
and meanwhile can effectively exclude the unimportant objective words.
In addition, the comparison further demonstrates the importance of the speaker information because of the better performance of our final model.

At the conversation-level, cross attention mechanism is used to identify closely-related utterances for a given utterance.
Here we show the indexes of the related utterances to study the learned information by MTL with SI.
As shown by the rightmost part of Figure \ref{figure_attention}, we can see that 
the cross attention mechanism can help to associate the utterances with the same speaker 
and the next utterances of the targeted speakers for a specific utterance.
For example, for the 9th utterance, the speaker is \textit{Phoebe},
and the targeted speaker is \textit{Joey}.
By MTL, the model connects the 7th and 11th utterances from the same speaker,
and meanwhile connects the 6th utterance from the target speaker. 
Intuitively, these utterances could be potential evidences to recognize the current emotion,
which is demonstrated by our final model.



\section{Conclusion}

In this work, we proposed a multi-task learning network for CER with the assistance of SI,
aiming to better capture speaker-related information, which has been demonstrated important for CER.
We exploited a strong baseline with BERT as backend, 
and then presented two neural network structures to bridge the two tasks for mutual interaction.
We conducted on two benchmark datasets to verify the effectiveness of the proposed method.
Results showed that our baseline is very strong, achieving the best performance compared with the previous state-of-the-art.
Further, the MTL based method can boost the performance significantly, 
leading to a new state-of-the-art in the literature.
Detailed experiments showed that our suggested components for MTL are both important.
In addition, we analyzed the proposed model in depth for comprehensive understanding.


\newpage
\bibliographystyle{named}
\bibliography{ijcai20}

\end{CJK}
\end{document}